\documentclass[10pt,twocolumn,letterpaper]{article}

\usepackage{wacv}
\usepackage{times}
\usepackage{epsfig}
\usepackage{graphicx}
\usepackage{amsmath}
\usepackage{amssymb}

\usepackage[table,xcdraw]{xcolor}
\usepackage{float}
\usepackage{multirow}
\usepackage{dblfloatfix}
\newcommand{\norm}[1]{\left\lVert#1\right\rVert}

\def\wacvPaperID{763} 

\wacvfinalcopy 

\ifwacvfinal
\def\assignedStartPage{9876} 
\fi

\ifwacvfinal
\usepackage[breaklinks=true,bookmarks=false]{hyperref}
\else
\usepackage[pagebackref=true,breaklinks=true,colorlinks,bookmarks=false]{hyperref}
\fi

\ifwacvfinal
\else
\fi

\begin{document}

\title{CoordiNet: uncertainty-aware pose regressor for reliable vehicle localization}

\author{Arthur Moreau\\
MINES ParisTech\\
Huawei Technologies\\
\and
Nathan Piasco\\
Huawei Technologies\\
\and
Dzmitry Tsishkou\\
Huawei Technologies\\
\and
Bogdan Stanciulescu\\
MINES ParisTech\\
\and
Arnaud de La Fortelle\\
MINES ParisTech\\
}

\maketitle

\begin{abstract}
    In this paper, we investigate visual-based camera relocalization with neural networks for robotics and autonomous vehicles applications. Our solution is a CNN-based algorithm which predicts camera pose (3D translation and 3D rotation) directly from a single image. It also provides an uncertainty estimate of the pose. Pose and uncertainty are learned together with a single loss function and are fused at test time with an EKF. Furthermore, we propose a new fully convolutional architecture, named CoordiNet, designed to embed some of the scene geometry. 

    Our framework outperforms comparable methods on the largest available benchmark, the Oxford RobotCar dataset, with an average error of 8 meters where previous best was 19 meters. We have also investigated the performance of our method on large scenes for real time (18 fps) vehicle localization. In this setup, structure-based methods require a large database, and we show that our proposal is a reliable alternative, achieving 29cm median error in a 1.9km loop in a busy urban area.
\end{abstract}

\section{INTRODUCTION}

Autonomous vehicle technologies need precise localization systems. These can be provided by several sensors (GPS, IMU, lidar, etc.), but Visual-Based Localization solution is becoming more and more reliable thanks to computer vision progress~\cite{piasco2018survey}. In particular, Deep Neural Networks are an appealing solution because of their low computation cost and reduced memory footprint compared to structure-based methods~\cite{Toft2020}, which currently provide the most accurate results. PoseNet~\cite{PoseNet} was the first "camera pose regressor": given images labelled by corresponding poses in a known environment, it learns to regress camera pose in an end-to-end way. These models can operate in real time for applications like robot localization or augmented reality.

However, poses predicted by learning are not as accurate as structure-based methods~\cite{Sattler2018}. In addition to making larger localization error, absolute pose regression generates outliers very far from the actual camera poses, so that trajectories are not consistent over time. These networks also fail to extrapolate outside of their training set boundaries and consequently struggle to outperform image retrieval baselines~\cite{Sattler2019}. Given a large and heterogeneous training dataset, CNNs are able to distinguish between static features useful for localization and transient observations depicted in the image (dynamic objects like vehicles and pedestrians, weather conditions or different illumination). But in practice such datasets are difficult to collect for localization tasks, resulting in an increased number of failures of pose regressors.


\begin{figure}[t]
    \begin{center}
        \includegraphics[width=\linewidth]{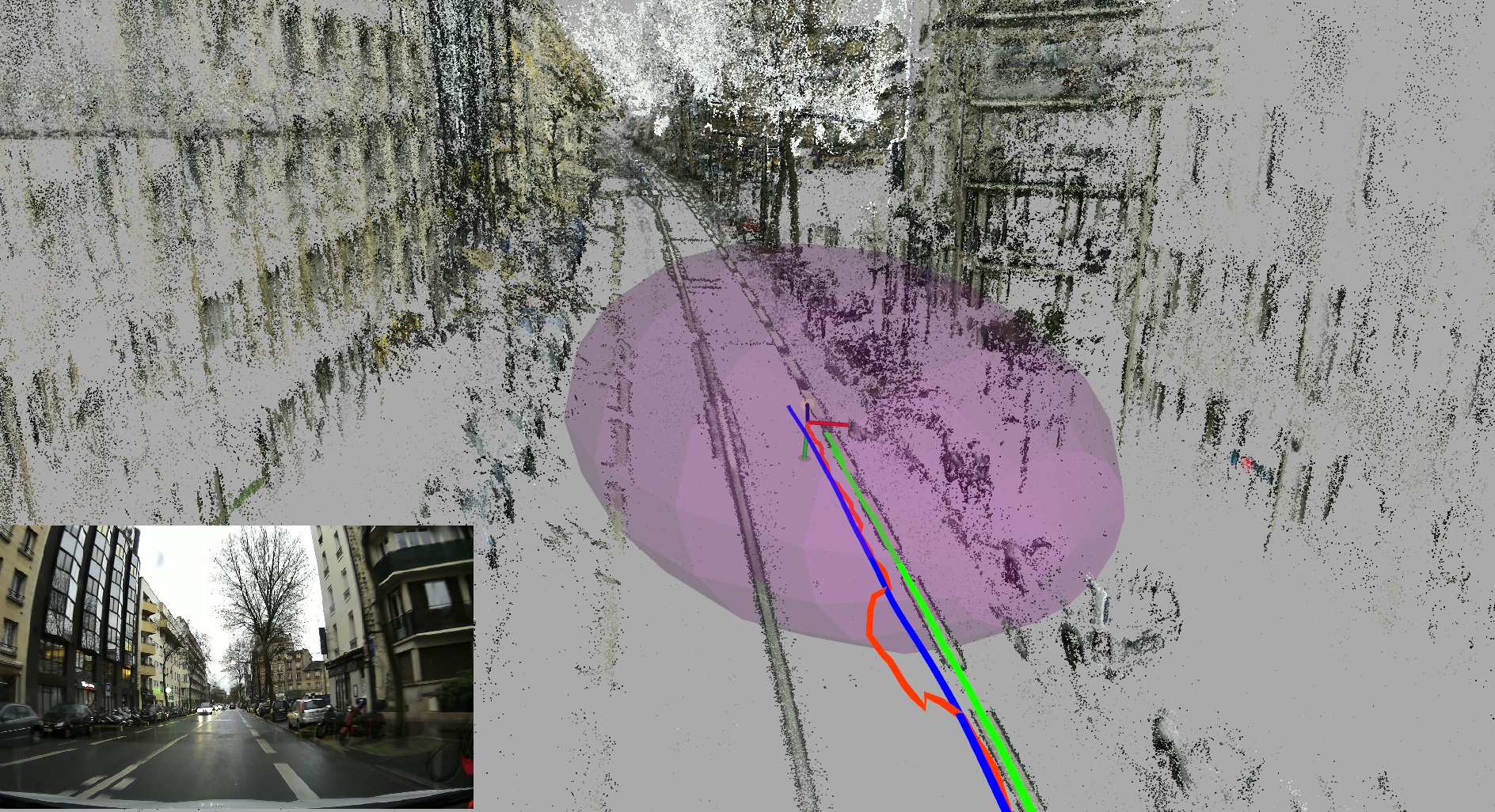}
    \end{center}
    \caption{\textbf{Visualization of CoordiNet online localization:} given an input camera image (bottom left), the network regresses 6-DoF poses (\textcolor{red}{red line}) with uncertainty estimate (\textcolor{purple}{purple ellipsoid}). These predictions can be fused with an EKF (\textcolor{blue}{blue line}), providing an accurate localization function ability (29cm median error to \textcolor{green}{ground truth}). Figure best viewed in color.}\label{fig:demo}
\end{figure}

\begin{figure*}[t!]
    \begin{center}
        \includegraphics[width=\linewidth]{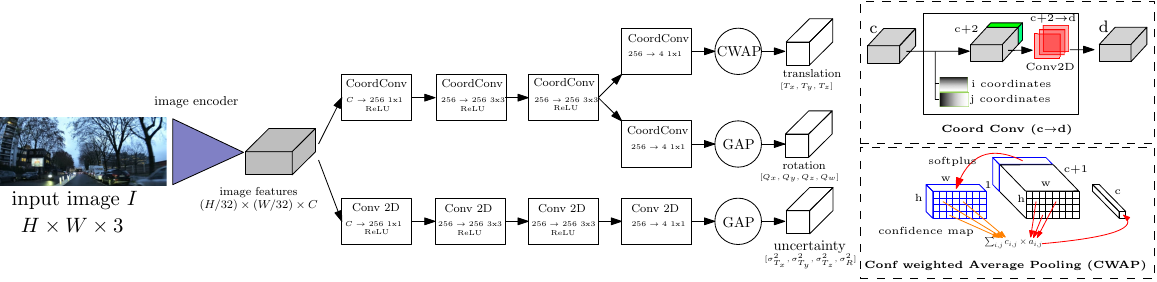}
    \end{center}
    \caption[fig:CoordiNet architecture]{\textbf{CoordiNet architecture:} the input image is sent to a pretrained image encoder, then pose and uncertainty are predicted from encoder features in two separate decoders. Pose decoder uses Coord convolutions layers. GAP and CWAP refer to Global Average Pooling and Confidence Weighed Average Pooling.}\label{coordinet}
\end{figure*}

In this paper, we propose a new pose regressor, named CoordiNet, designed to mitigate some of the aforementionned limitations.

Even if visual localization is considered as a geometric task (2D to 3D matches between 2D image features and 3D points in the environment), pose regressors do not solve it using geometric reasoning~\cite{Sattler2019}. CNNs are widely known for their ability to extract relevant features from images, but surprisingly they fail to solve trivial geometric tasks like retrieving the cartesian coordinate of a colored pixel in a white image~\cite{coordconv}. Our work builds up on the previous idea and embed geometrical hints in the architecture of camera pose regression CNN. Our proposed architecture takes advantage of the Cartesian representation of the image with the use of Coord Convolutions~\cite{coordconv} and spatial self-attention awareness thanks to a confidence-weighted average pooling~\cite{fc4} which replaces the standard global average pooling.

In order to make a robust localization system, one needs to know when the predicted pose is not reliable. Computing uncertainty coupled with pose regression is a common way to handle this problem~\cite{bayesianposenet}. However, widely used approaches to compute uncertainty for pose regression have limitations for practical applications. They generate multiple hypothesis for each single image at inference time, and then compute mean and variance to estimate pose and uncertainty~\cite{bayesianposenet,RVL}. This increases a lot computational complexity because several inferences are required. For practical applications, one needs uncertainty to be estimated jointly with the pose regression and such estimated uncertainty should be highly correlated to a potential level of errors of regressed poses. To address this requirement, CoordiNet was designed to predict uncertainty from activations of an hidden layer, so it can be jointly computed at inference with pose regression and learn to associate potential failures with content of input image.

Both poses and uncertainty are learned together with a unified loss function. Finally, the predicted uncertainties can be used in any post processing step as a covariance matrix attached to the pose prediction, for instance in a graph factor~\cite{Dellaert2017} formulation or within a Kalman filter.

The main contributions of this paper are:
\begin{itemize}
    \item a method to train jointly pose prediction and uncertainty, with reliable uncertainty estimate and an improved training stability,
    \item a new fully convolutional architecture that integrates geometric clues and outperforms monocular state of the art methods on all public benchmarks with a large margin,
    \item an extensive evaluation of deep pose regressors in several areas with large scale datasets, showing that CoordiNet can be used in real-time (18 Hz with ROS implementation on a RTX2080 embedded GPU) for vehicle localization,
    \item we show that pose predictions combined with reliable uncertainties in a simple EKF exhibits smooth trajectories and remove outliers.
\end{itemize}

In the following section we review the related work. In section \ref{sec:method}, we describe our architecture and the intuition behind it. Later, we present how to learn pose and uncertainty together. In section~\ref{sec:results}, we present results of multiple experiments to prove the benefit of our approach: ablation study, comparison with related methods and experiments on large-scale applications. Section~\ref{sec:conclusion} concludes the paper.

\section{RELATED WORK}
\label{sec:related_work}

In this section we present current status of research concerning camera pose regression with neural networks and uncertainty estimation for these methods. We also briefly discuss the benefit of this approach compared to other visual-based localization solutions.

PoseNet~\cite{PoseNet} was the first deep learning method to regress absolute camera pose from an image in an end-to-end way. It uses a pretrained image encoder (GoogLeNet in the first version), followed by a global average pooling and 2 fully connected layers which output 7 scalars (3 for translation and 4 for rotation). It works in real time but has several drawbacks. We present here these limitations and how they have been addressed by previous work.

State-of-the-art image based localization methods are built on a rich representation of the environment: a 3D point cloud where each 3D point is linked to a set of handcrafted or learned descriptors~\cite{sarlin2019coarse}. New image poses are computed by matching 2D features in the image to 3D points in the point cloud (F2P). Pose regressors have the main advantage of carrying a much more compact scene representation (weights of the network) compared to standard image localization methods (3d point cloud or image database). Meanwhile, overall accuracy of these networks is one step lower than other visual-based localization solutions. In order to increase pose accuracy, loss function have been improved in the second version of PoseNet~\cite{geometric_loss_function}. It uses homoscedatic uncertainty to weight translation and rotation loss in order to equilibrate the training. Several innovations have been proposed to improve the architecture of the network: Hourglass network~\cite{Hourglass} proposes an encoder-decoder architecture, AtLoc~\cite{AtLoc} and RVL~\cite{RVL} use an attention module before the fully-connected layers. VLocNet++~\cite{VLocNet++} investigates multi-task learning to improve pose accuracy and learn to predict pose and semantic segmentation with a shared backbone. Another type of pose regressors produce a dense map of scene coordinates where each image pixel is associated to a 3D point in the scene~\cite{Brachmann_2018_CVPR}. The pose can be computed using at least 3 of these predicted scene coordinates. These type of networks can be more accurate than F2P solutions but are not able to cover large areas~\cite{Brachmann_2019_ICCV}.

Then, inference on consecutive images in a video results in a discontinuous sequence of pose (see figure~\ref{fig:demo}). This can be improved by providing previous images as an additional input. Then, \cite{lstmpose} and Vidloc~\cite{vidloc} propose to use a LSTM architecture to handle temporal processing. VLocNet~\cite{VLocNet} is trained with an additional constraint on the relative pose of consecutive images to improve accuracy and provide smooth pose sequences. MapNet~\cite{mapnet2018} proposes to constraint relative poses with signals from other sensors like visual odometry or GPS. \textit{Xue et al.}~\cite{graph_neural_network} model the problem with a graph neural network which learn the dependency between 8 frames. In our work, we rely on an external filtering tool (EKF) to post-process poses and generate a smooth trajectory. Similar post-processing method, with a pose graph optimization formulation, have already been used in~\cite{mapnet2018,RVL}.

Finally, as we predict absolute pose, localization errors can be very large when failure cases happen. To handle this, we need a way to filter out these outliers. Uncertainty estimation is an ideal tool because it can be used to post process poses according to uncertainty values, using tools like Kalman filters~\cite{zhou2020kfnet} or factor graphs~\cite{Dellaert2017}. In~\cite{whatuncertainties}, the authors provide a very clear study about uncertainty in computer vision with Bayesian Deep Learning. There are 2 types of uncertainty one can model:
\begin{itemize}
    \item Epistemic uncertainty (or model uncertainty) is the uncertainty on the weights of the network. It measures the degree of knowledge of the network about input data. A simple way to approximate it is Monte Carlo Dropout~\cite{gal2016dropout} (MCD): train a network with dropout layers and keep them active at inference time. Several inference on the same data will provide a sample of results. The mean is used as prediction and the variance is epistemic uncertainty. Bayesian PoseNet~\cite{bayesianposenet} uses this method to estimate uncertainty of pose regression. RVL~\cite{RVL} proposes a prior guided dropout on input image: it removes areas where dynamic objects appear and allows to generate a Monte Carlo sample too.
    \item Aleatoric uncertainty (or data uncertainty) is the variance of the network output which can be caused by noise in input data. We can choose between heteroscedatic aleatoric uncertainty which can vary with input data and homoscedatic uncertainty which measures the overall variance in the outputs of a task as it is not dependent on input data. Heteroscedatic can be predicted directly from the network, as shown in \cite{lakshminarayanan2017simple,whatuncertainties}, by using a maximum likelihood loss function. This formulation can be extended to multivariate uncertainty, as shown by~\cite{multivariate} where correlation between outputs variables is learned. Finally, HydraNet~\cite{peretroukhin2019probabilistic} combines epistemic and aleatoric uncertainties to provide consistent orientation estimates.
\end{itemize}
We decide to learn heteroscedatic uncertainty as an auxiliary task during the training of our model. Thus, we obtain an uncertainty estimate at test time for less computation than with MCD.  

\section{METHOD}
\label{sec:method}

In this section, we first present our network architecture. Then we show how pose and uncertainty estimate are learned together during training and finally present the fusion at test time for a reliable localization.

\subsection{Adapting convolutional architectures to exploit spatial information}

\begin{figure}[h]
\label{pooling_activations}
   \includegraphics[width=\linewidth]{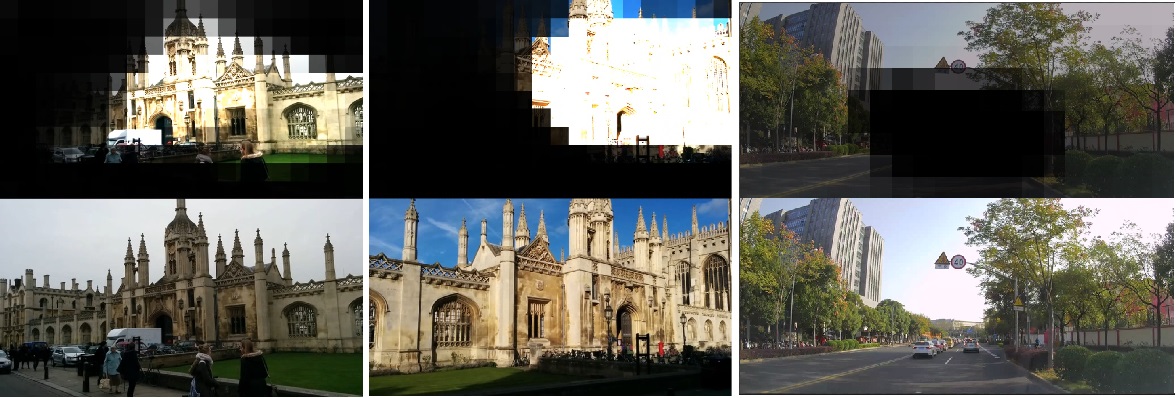}
   \caption{\textbf{Visualization of pooling activations:} bottom images are inputs of the model, top images are inputs multiplied by the upsampled confidence map of the pooling layer (presented in Figure~\ref{coordinet}).}
\end{figure}

Inspired by Features-to-Points approaches~\cite{hartley2003multiple}, where location of 2D features in the image is linked to 3D points in the environment, we designed our architecture to follow this approach. CNNs are already good to retrieve the topological-level position~\cite{Piasco2020} (\textit{e.g.} at a street level) thanks to image content, but in order to reach higher accuracy, one could compare the precise location of objects of interest in image coordinates. Current architectures are inspired from image classification task which deals with semantics and is expected to be invariant to spatial location of objects in the image. However, as our task deals with scene geometry, we want to mitigate this translation invariance in order to take account of precise feature positions.

To do so, we replace standard 2D convolution by Coord convolution, introduced by \cite{coordconv} in a study about CNN limitations on simple geometric tasks. Coord convolutions concatenate 2 additional channels, that contain hard-coded pixel coordinates, to the input tensor before applying the convolution, as illustrated in Figure~\ref{coordinet}.

Finally, we propose to use confidence weighted average pooling (CWAP), instead of global average pooling (GAP), inspired from previous success of this method in other applications~\cite{fc4}. In order to transform a feature map into a single scalar, GAP simply computes the mean of the feature map. CWAP computes a weighted mean using an additional channel as a confidence map, providing a weight for each spatial location. These weights are predicted according to previous layer activation so we can compare this computation to a low-cost self-attention mechanism. Examples of activation mask of the confidence map are shown in figure~\ref{pooling_activations}. We observe that on small scenes of Cambridge Landmarks, the pooling always highlights the same object regardless of camera pose (here front of Kings College). On larger scenes, there is no common object visible in all the scene. In this case, the pooling masks areas where dynamic objects appear.

The entire architecture of our model is depicted in Figure~\ref{coordinet}. We use two decoder heads to predict poses and uncertainties from a latent representation obtained by an image encoder. Our architecture is fully-convolutional, \textit{i.e} the number of parameters of the decoder does not depend on the size of the input image. Compared to a standard pose regressor that uses fully connected layers to regress the final pose, our decoder contains one order of magnitude less parameters (\textit{e.g.} for an image size of $360 \times 640$ and a latent representation with $512$ feature maps, a one-layer fully connected decoder has $0.8M$ parameters compared to $0.6M$ parameters in our pose decoder).

\subsection{Joint learning of pose and heteroscedatic uncertainty}
\label{jointlearning}

Pose regression problem can be modeled by the equation
$Y = f_W(I)$,
where $f$ is the neural network with weights $W$, $Y$ is the predicted pose and $I$ is the input image. In Bayesian Deep Learning~\cite{whatuncertainties}, $W$, $I$ and $Y$ are considered as random variables and uncertainty is the variance of these variables.

In our problem, $Y \in SE(3)$. We decompose the 6-DoF pose with a 3D translation vector $[T_{x},T_{y},T_{z}]$ and a unit quaternion $[Q_x,Q_y,Q_z,Q_w]$ for rotation.

For translation, we model $T_{x},T_{y},T_{z}$ as 3 independent Gaussian variables centered on the actual pose with variances $\sigma^2_{T_{x}},\sigma^2_{T_{y}},\sigma^2_{T_{z}}$. Our framework predicts these $\sigma^2$ uncertainties, which represent the expected noise in the output pose, depending on input image. As we made it data dependent, it is called heteroscedatic uncertainty. In practice, each uncertainty is learned relatively to a loss function. In our case, we want to have one uncertainty estimate for each translation component, so we use 3 separate $L_{1}$ translation losses $L_{T_x}$, $L_{T_y}$, $L_{T_z}$.

For rotation, we can not use the same formulation, first because individual components of a unit quaternion are clearly not independent, but also because the 3D rotation group $SO(3)$ is not euclidean. As a result, we optimize rotation with a single loss function $L_R$, resulting in a single rotation uncertainty estimate $\sigma^2_{R}$. An other manner to learn uncertainty for rotation have been proposed in~\cite{peretroukhin2019probabilistic}, but we found out that in practice our 1-dimension uncertainty estimation work well and leave the integration of multivariate rotation uncertainty estimation in pose regression for future work. Our choice for $L_R$ is the geodesic distance between rotations, defined as the minimal angular difference between 2 rotations:

\begin{equation}
L_R = cos^{-1}((tr(M_{pred} M_{GT}^{-1}) - 1)/2)
\end{equation}
where $M_{pred}$ and $M_{GT}$ are predicted and reference 3D rotations, converted to rotation matrices. \cite{rotation_continuity} has shown that this optimization objective performs better than $L_2$ loss.

Finally, we combine these 4 loss functions by minimizing the negative log-likelihood of our model. We do not learn $\sigma^2$ directly but use $s = \log  \sigma^2$ for numerical stability, following \cite{geometric_loss_function}. Our final optimization objective becomes:
\begin{equation}
L_{\sigma}(I) = \sum_{i \in [T_x, T_y, T_z, R]} L_{i}(I)  e^{-s_{i}(I)} + s_{i}(I)
\label{eq:NLL}
\end{equation}

This loss function is actually the same than the \textit{learnable weights pose loss}~\cite{geometric_loss_function}, except that uncertainty values are outputs of the network instead of free scalar values (homoscedatic uncertainty).

To minimize this loss function, the network needs to learn accurate poses in order to decrease $L_i$ losses. When a challenging image is provided, the network can predict high uncertainties in order to reduce the weights of regression losses in the objective. The second term acts as a penalization term to avoid infinite uncertainties. The best way to minimize this cost is to predict uncertainties proportional to loss values. 

This method also has desirable effects for training. When used with homoscedatic uncertainty, each individual loss will contribute to the final loss with approximately the same weight. As our uncertainties vary with input data, this property is extended at a data level : each training sample will contribute to the batch loss with an approximately equal weight, whereas usually large errors contribute more. 
Contrary to the object classification problem where samples having larger error should be main target for optimizing the network, we want every sample to be considered equally important in the optimization process.

\subsection{Localization under uncertainty}
\label{locunderuncertainty}

At test time, we fuse together the regressed pose with learned uncertainties in order to filter out failure cases and obtain a smooth and temporally consistent trajectory. This is a desirable property in autonomous driving and robotics applications, because localization could be directly used by the planning algorithm to compute control command.

We propose to use an Extended Kalman Filter (EKF) with an omnidirectional motion model for this fusion step. Integration is done by providing only the absolute pose measurement given by the network to the filter. We attach a simplified diagonal covariance matrix $\Sigma$ to each measurement, defined by:

$$
\Sigma = I_{6} * 
\left[
\sigma^2_{T_{x}} 
\sigma^2_{T_{y}} 
\sigma^2_{T_{z}} 
\sigma^2_{R} 
\sigma^2_{R} 
\sigma^2_{R} \\ 
\right]^t.
$$

This formulation is limited to represent uncertainty in $SE(3)$: first because in practice the covariance between variables can be non-zero. Another limitation discussed earlier is the use of a one dimensional rotation uncertainty $\sigma^2_{R}$.
We tried to use more sophisticated formulations, where non-diagonal coefficients are learned, inspired by \cite{multivariate,peretroukhin2019probabilistic}, but observed a lower pose regression accuracy with this formulation. One reason could be that vectors in equation~\ref{eq:NLL} are replaced by matrices, leading to a lower numerical stability during training. We plan to improve our model of covariance matrix in $SE(3)$ as a future work. However, we show in \ref{loc_uncertainty_exp} that our proposal is sufficient in practice to reach our target: a consistent trajectory where outliers are filtered.

\begin{figure}[t]
\centering

\includegraphics[width=\linewidth]{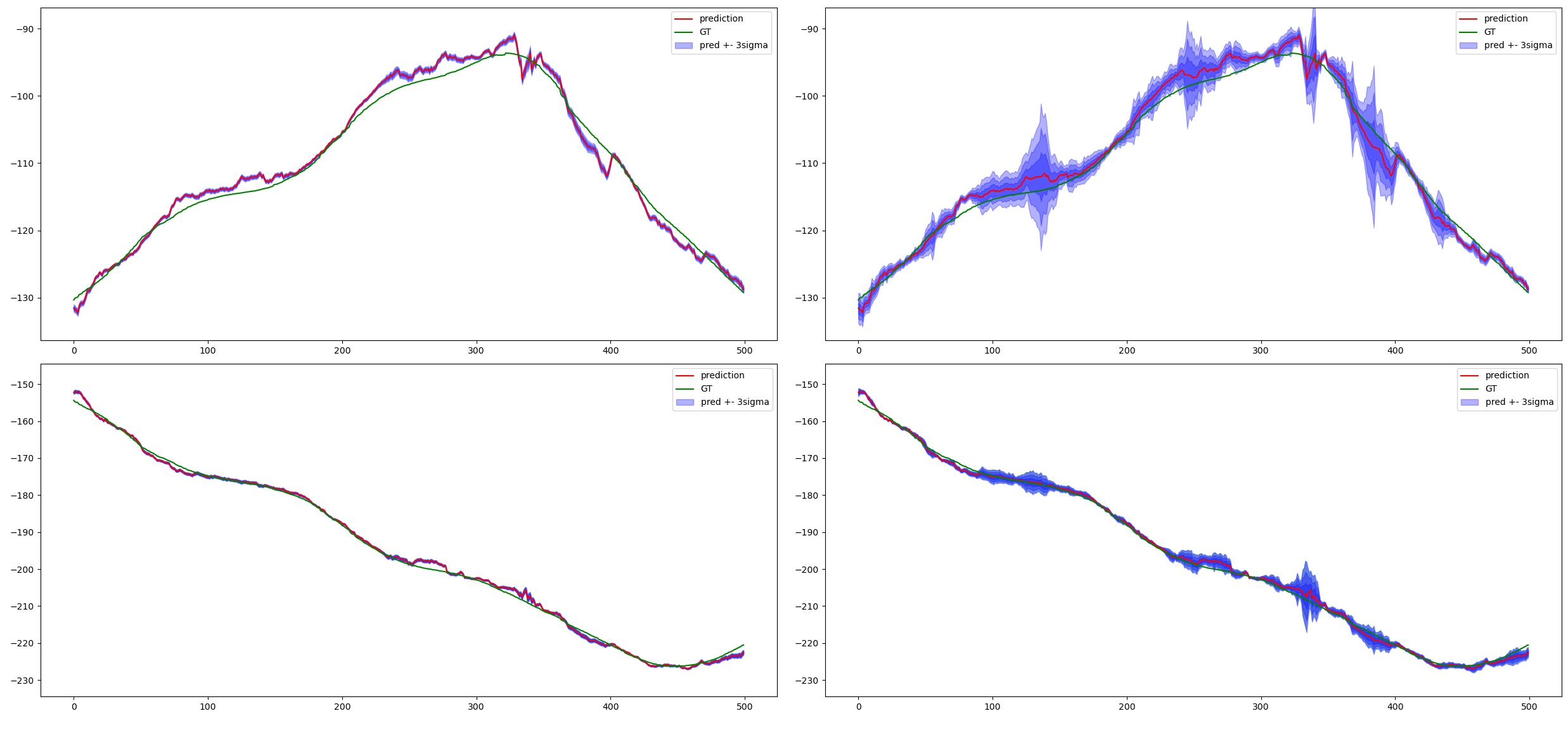}
   
\caption{\textbf{Calibrated uncertainties.} Comparison between uncalibrated  (left) and calibrated (right) uncertainties, plotted for x (top) and y (bottom) axis.}\label{calibrated_uncertainty}
\end{figure}

\label{subsec:calibrated_covariance}
\textbf{Uncertainty calibration:} During the evaluation of our method, we observed that learned uncertainties often underestimate the actual error (see \ref{loc_uncertainty_exp}). This is caused by overfitting: at the end of the training procedure, the model performs very well on training images and the uncertainty layer learns a distribution of errors which does not represent the actual distribution in test conditions. To mitigate this effect, we propose a 2 steps training procedure: available training data is split in a training set and a validation/calibration set. We first train CoordiNet with the training set with the procedure described in \ref{jointlearning}, and then fine-tune the uncertainty layer on the calibration set while all other layers are freezed. This enable to calibrate uncertainties on examples representative of test conditions. Since the use of a validation set is a common machine learning practice, this calibration step does not make our method more data-intensive than others. We show in figure~\ref{calibrated_uncertainty} the benefit of uncertainty calibration for proper uncertainty estimation on a test sequence.

\begin{table*}[!t]
\footnotesize
\centering

\caption{\label{tab:oxford_results}Results on Oxford RobotCar dataset (Camera pose regression without post-processing). CoordiNet is compared to other neural network pose regressors. Green cells represent best translation results. Orange cells represent previous best monocular method, to whom CoordiNet should be compared for a fair evaluation. All methods use ResNet34 as image encoder. Table best viewed in color.}

\vspace{0.4cm}

\begin{tabular}{|l|l|l|l|l|l|l|l|l|}
\hline
\multicolumn{3}{|c|}{\textbf{Oxford RobotCar Dataset}} & \multicolumn{3}{c|}{\textbf{Mean error comparison}}                                                                    & \multicolumn{3}{c|}{\textbf{Median error comparison}}                                                 \\ \hline
\textbf{Method}      & \textbf{Loss}  & \textbf{input}  & \textbf{Loop}                         & \textbf{Full}                         & \textbf{Average}                       & \textbf{Loop}                 & \textbf{Full}                 & \textbf{Average}                      \\ \hline
\textbf{CoordiNet}   & Heterosc.          & 1               & 4.15m / 1.44° & 14.96m / 5.74° & 9.56m / 3.59° & \cellcolor[HTML]{92D050}2.27m / 0.86° & \cellcolor[HTML]{92D050}3.55m / 1.14° & \cellcolor[HTML]{92D050}2.91m / 1.00°         \\ \hline
\textbf{CoordiNet}   & Homosc.            & 1               & \cellcolor[HTML]{92D050}4.06m / 1.44° & \cellcolor[HTML]{92D050}11.99m / 6.15°  & \cellcolor[HTML]{92D050}8.03m / 3.80° & 2.42m / 0.88° & 4.21m / 1.06°                         & 3.32m / 0.97°                                \\ \hline
AD-MapNet            & Homosc.            & 2               & 6.45m / 2.98°                         & 19.18m / 4.60°                        & 12.82m / 3.79°                         &                               &                               &                                       \\ \hline
AtLoc+               & Homosc.            & 2               & 7.53m / 3.61°                         & 21.0m / 6.15°                         & 14.27m / 4.88°                         & 4.06m / 1.98°                 & 6.40m / 1.50°                 & 5.23m / 1.74°                         \\ \hline
AtLoc                & Homosc.            & 1               & 8.73m / 4.63°                         & \cellcolor[HTML]{FFC000}29.6m / 12.4° & \cellcolor[HTML]{FFC000}19.17m / 8.52° & \cellcolor[HTML]{FFC000}5.36m / 2.10°                 & \cellcolor[HTML]{FFC000}11.1m / 5.28°                 & \cellcolor[HTML]{FFC000}8.23m / 3.69° \\ \hline
AD-PoseNet         & Homosc.            & 1               & \cellcolor[HTML]{FFC000}6.40m / 3.09° & 33.82m / 6.77°                        & 20.11m / 4.93°                         &                               &                               &                                       \\ \hline
MapNet               & Homosc.            & 2               & 9.29m / 3.34°                         & 44.61m / 10.38°                       & 26.95m / 6.86°                         &                               &                               &                                       \\ \hline
PoseNet              & Homosc.            & 1               & 7.9m / 3.53°                          & 46.61m / 10.45°                       & 27.26m / 6.99°                         &                               &                               &                                       \\ \hline
\end{tabular}

\vspace{1mm}
\end{table*}

\begin{table*}[!t]
\footnotesize
\centering

\caption{\label{tab:cambridge_results}Results on Cambridge Landmarks dataset. Refer to legend of table~\ref{tab:oxford_results} for color meaning.}

\vspace{0.4cm}

\begin{tabular}{|l|l|l|l|l|l|l|l|}
\hline
\textbf{Method}    & \textbf{Backbone} & \textbf{input} & \multicolumn{1}{c|}{\textbf{Old Hospital}} & \textbf{Kings College}                & \textbf{StMarysChurch}                & \textbf{Shop Facade}                  & \multicolumn{1}{c|}{\textbf{Average}} \\ \hline
VLocNet            & ResNet50          & 2              & 1.07m / 2.41°                              & 0.84m / 1.42°                         & \cellcolor[HTML]{92D050}0.63m / 3.91° & \cellcolor[HTML]{92D050}0.59m / 3.53° & \cellcolor[HTML]{92D050}0.78m / 2.82° \\ \hline
\rowcolor[HTML]{E7E6E6} 
\textbf{CoordiNet} & EffNet b3         & 1              & \cellcolor[HTML]{92D050}0.97m / 2.08°      & \cellcolor[HTML]{92D050}0.70m / 0.92° & 1.32m / 3.56° & 0.69m / 3.74°  & \cellcolor[HTML]{92D050}0.92m / 2.58° \\ \hline
\rowcolor[HTML]{E7E6E6} 
\textbf{CoordiNet} & ResNet34          & 1              & 1.43m / 2.86° & 0.80m / 1.22°                                & 1.32m / 4.10°                                 & 0.73m / 4.69°                                 &  1.07m / 3.22°                                      \\ \hline
LSTM-Pose          &                   & 2              & 1.51m / 4.29°                              & 0.99m / 3.65°                         & 1.52m / 6.68°                         & 1.18m / 7.44°                         & 1.30m / 5.51°                         \\ \hline
PoseNet            & ResNet34          & 1              & 2.17m / 2.94°                              & 0.99m / 1.06°                         & 1.49m / 3.43°                         & 1.05m / 3.97°                         & 1.43m / 2.85°                         \\ \hline
AD-PoseNet         & ResNet34          & 1              & \multicolumn{1}{c|}{non reported}          & 1.3m / 1.67°                          & 2.28m / 4.80°                         & 1.22m / 4.64°                         & /                                     \\ \hline
MapNet             & ResNet34          & 2              & 1.94m / 3.91°                              & 1.07m / 1.89°                         & 2.00m / 4.53°                         & 1.63m / 4.22°                         & 1.66m / 3.64°                         \\ \hline
Bay. PoseNet       &                   & 1              & 2.57m / 5.14°                              & 1.74m / 4.06°                         & 2.11m / 8.38°                         & 1.25m / 7.54°                         & 1.92m / 6.28°                         \\ \hline
\end{tabular}
\end{table*}

\section{EXPERIMENTS}
\label{sec:results}

In this section, we evaluate CoordiNet in multiple scenarios. First, we compare to related methods on public datasets. Next, we investigate how performance of CoordiNet scales with the size of the dataset by focusing on privately collected datasets which are orders of magnitude larger than public ones. Third, we demonstrate that once CoordiNet is fused with EKF it could be considered as a good alternative for reliable localization in practical tasks. Finally, we conduct an ablation study to highlight importance of individual contributions we propose in this paper.

\subsection{Results on public benchmark}

\subsubsection{Oxford Robotcar}

First, we compare our method to related work on Oxford Robotcar dataset~\cite{RobotCarDatasetIJRR}. We reproduced the experiments done by Huang et al.~\cite{RVL}. The model is trained on 2 scenes (Full and Loop) using only 2 training sequences in each case. While Full is tested on one sequence, 2 sequences are used for Loop. Full scene is particularly challenging because the area is very large (9 kms) and the network can fail to generalize with only 2 different sequences in the training set. We resize input images to $256 \times 455$. We train our model with ResNet34 encoder for a fair comparison with previous methods, and keep a fixed learning rate of $1e^{-4}$. We compare CoordiNet to other monocular methods, but also to sequential methods using several images as input. Unlike other benchmarks, comparison on Oxford dataset is usually done comparing mean error instead of median error. Our opinion is that median error is more meaningful because it is not corrupted by outliers, so we report both.

\begin{figure}[!b]
   \includegraphics[width=\linewidth]{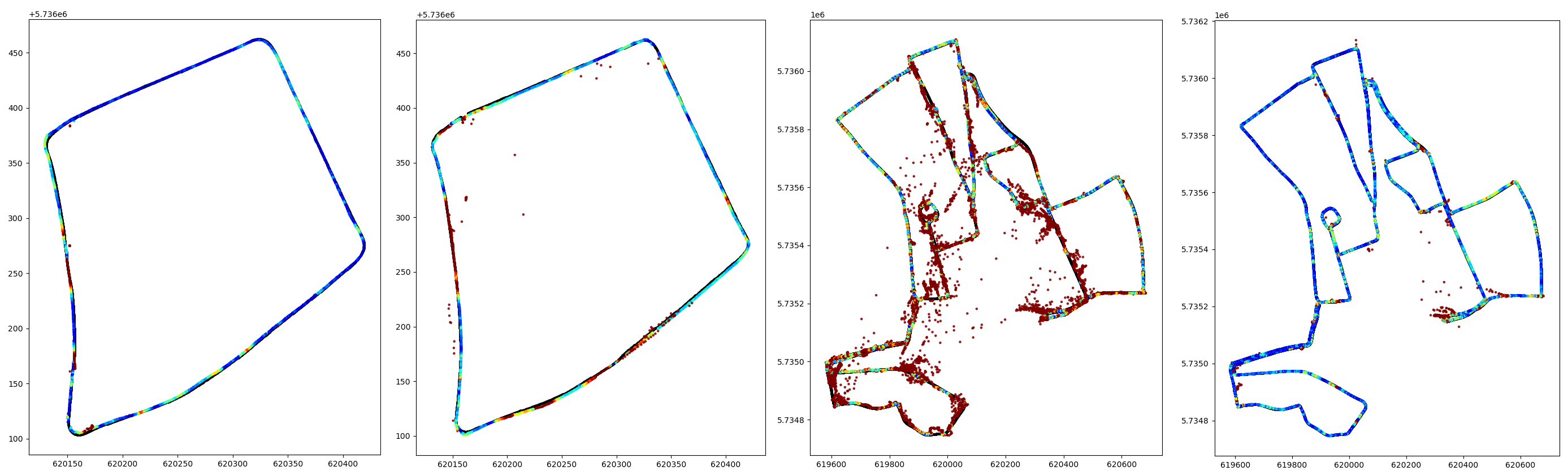}
   \caption{Results on Oxford experiment. Left images are Loop results (2014-06-23-15-36-04, 2014-06-26-08-53-56) trained with 2 sequences, right images are Full results (2014-12-09-13-21-02). Middle right is trained with 2 sequences, right is trained with 15. Color map represents errors at a given location: blue is $\sim 1m$ error and red is $>5m$ error. Figure best viewed in color.}\label{oxford_trajs}
\end{figure}

Results are reported in table~\ref{tab:oxford_results} and figure~\ref{oxford_trajs}. Not only CoordiNet beats other monocular methods but also sequential networks with a margin of 4 meters in translations. Rotation is also improved on Loop scene and performs equally well as sequential methods on Full scene. We observe that training with uncertainty improves median accuracy by 13\% even if mean error is higher. It confirms that this training method attenuates the impact of outliers.

\subsubsection{Cambridge Landmarks}

Then, we also report CoordiNet performances on Cambridge Landmarks. This dataset contains several small outdoor scenes with small training datasets. We train our model with both ResNet34 and EfficientNet-b3, and again compare CoordiNet to monocular and sequential methods. Images are computed at full resolution ($640\times350$). Results are reported in table~\ref{tab:cambridge_results}. Compared to other monocular methods, CoordiNet reports best results on all scenes.

\begin{figure*}[!b]
\centering
\includegraphics[width=0.48\linewidth]{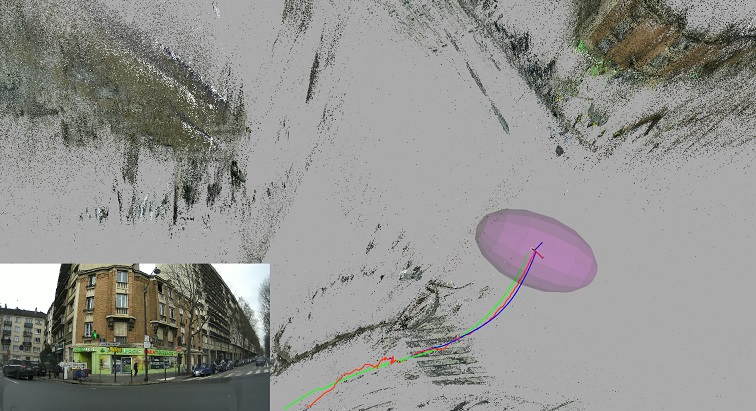}
\includegraphics[width=0.48\linewidth]{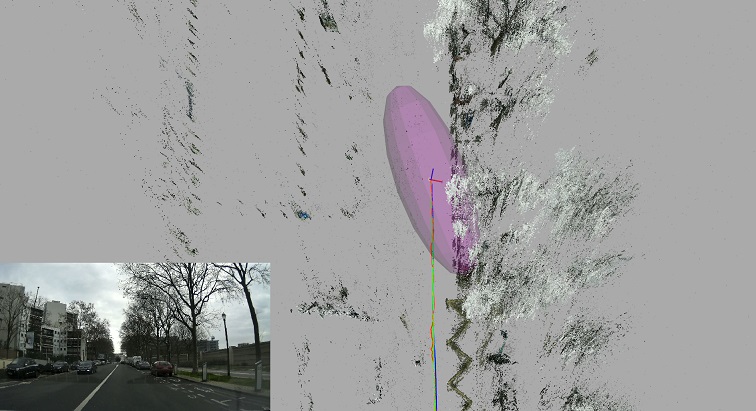} \\
\vspace{1mm}
\includegraphics[width=0.48\linewidth]{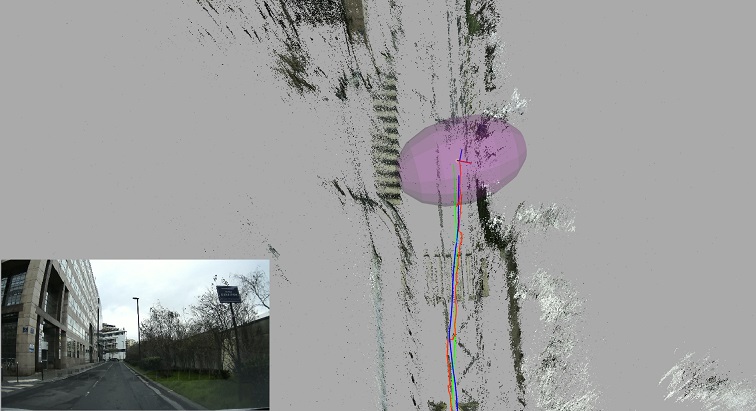}
\includegraphics[width=0.48\linewidth]{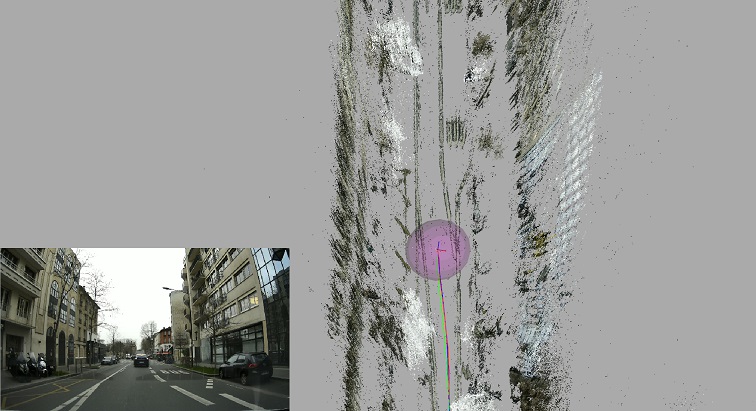}
   \caption{\textbf{CoordiNet and EKF trajectories:} CoordiNet sequences of poses (\textcolor{red}{red line}) are shown with the uncertainty estimate of the current pose (\textcolor{purple}{purple ellipsoid}). EKF trajectory (\textcolor{blue}{blue line}) and ground truth (\textcolor{green}{green line}) are also displayed. Figure best viewed in color.}\label{screenshots}
\end{figure*}

\subsection{Results on large-scale datasets}
\label{large_scale_experiments}

\begin{figure}[ht]
\centering

\includegraphics[width=0.33\linewidth]{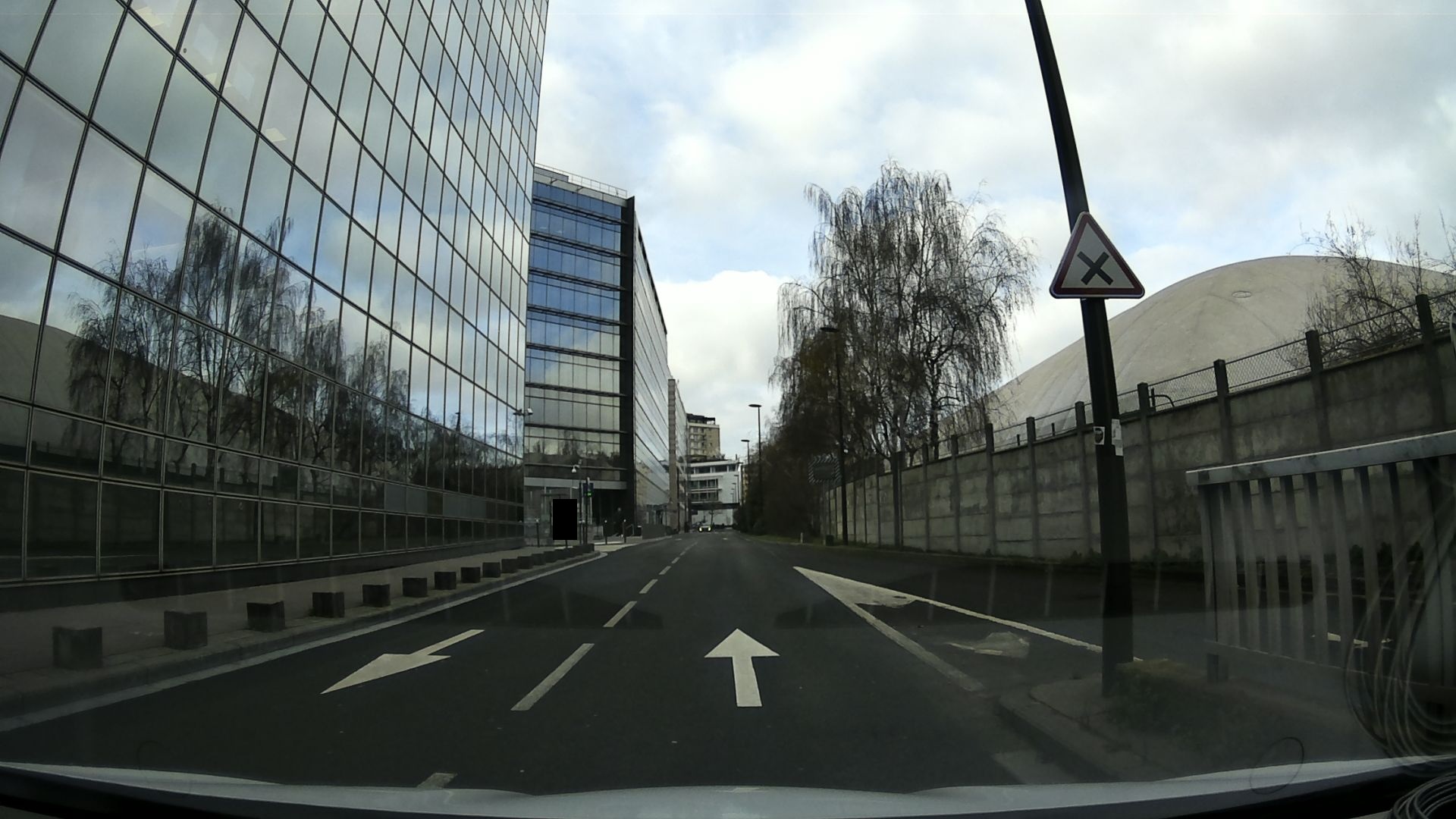}\includegraphics[width=0.33\linewidth]{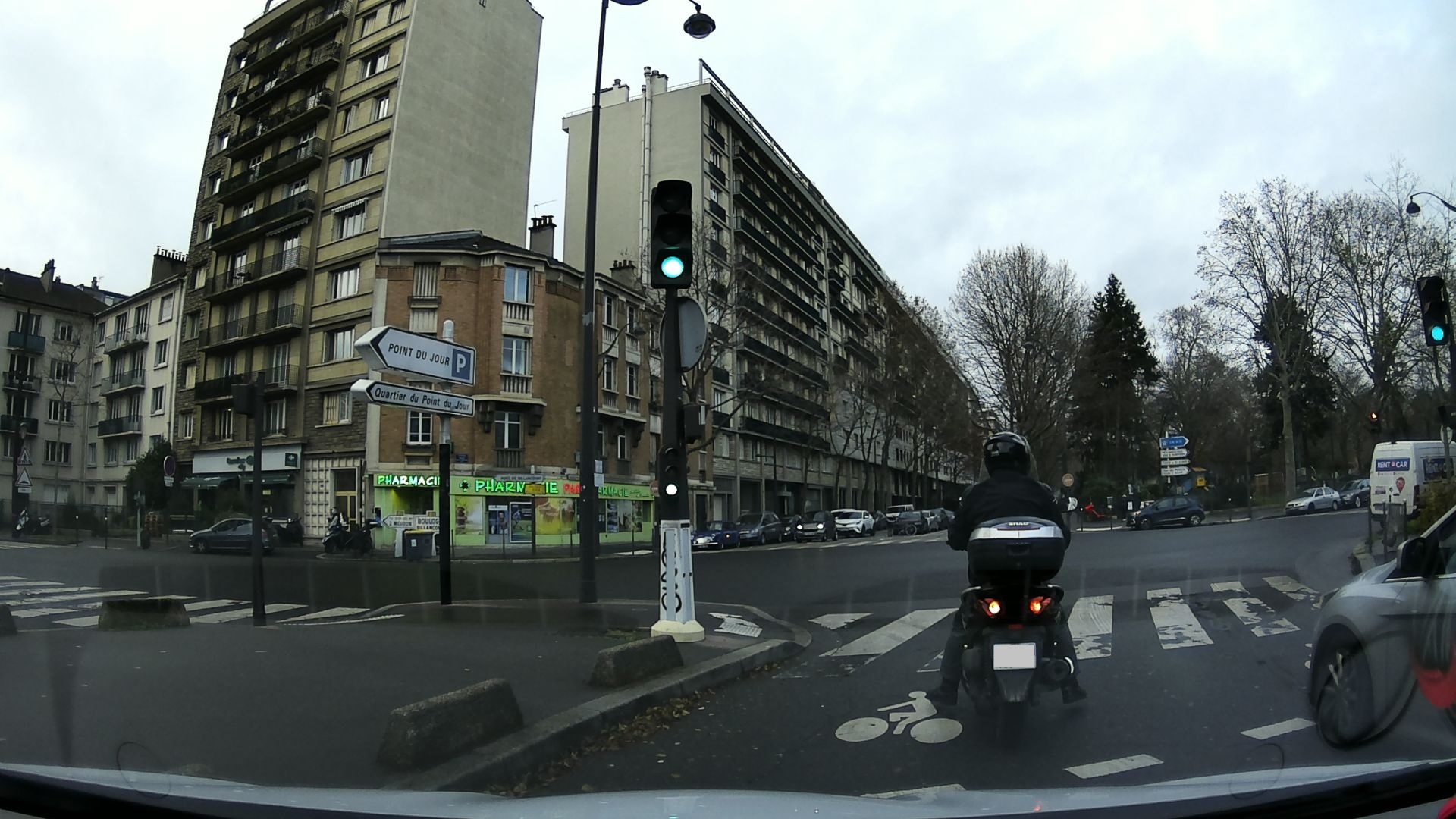}\includegraphics[width=0.33\linewidth]{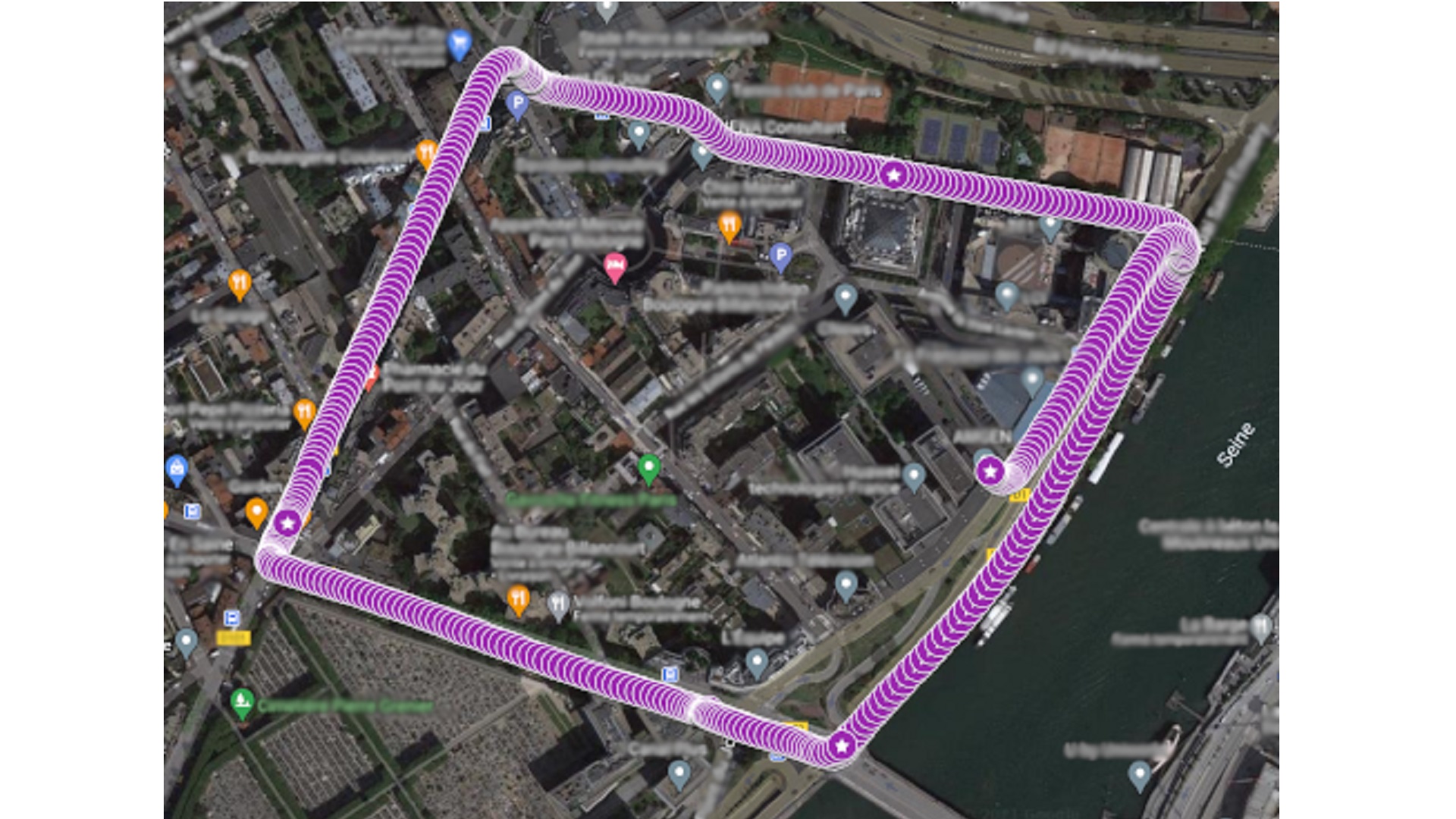}\\

\includegraphics[width=0.33\linewidth]{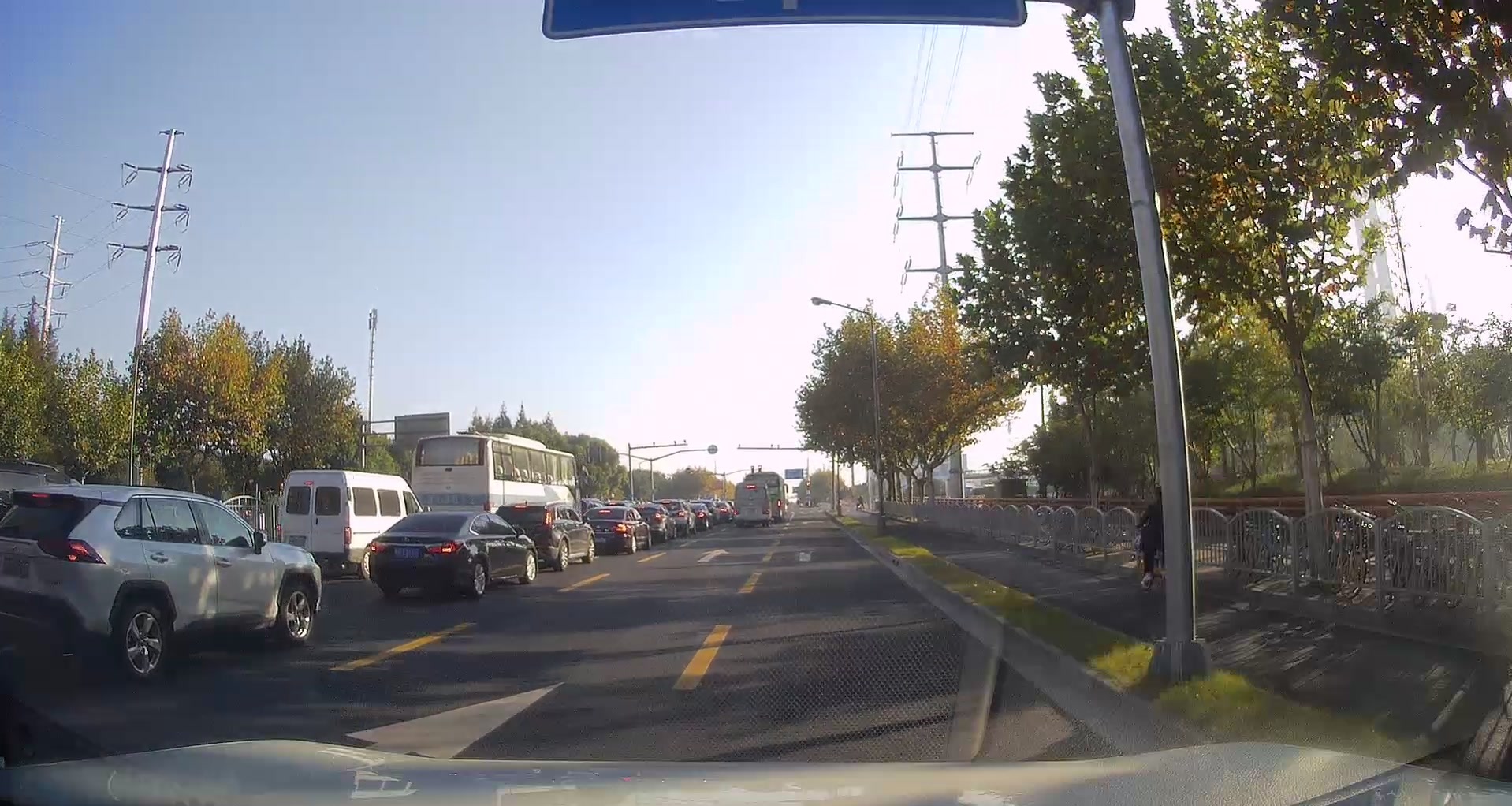}\includegraphics[width=0.33\linewidth]{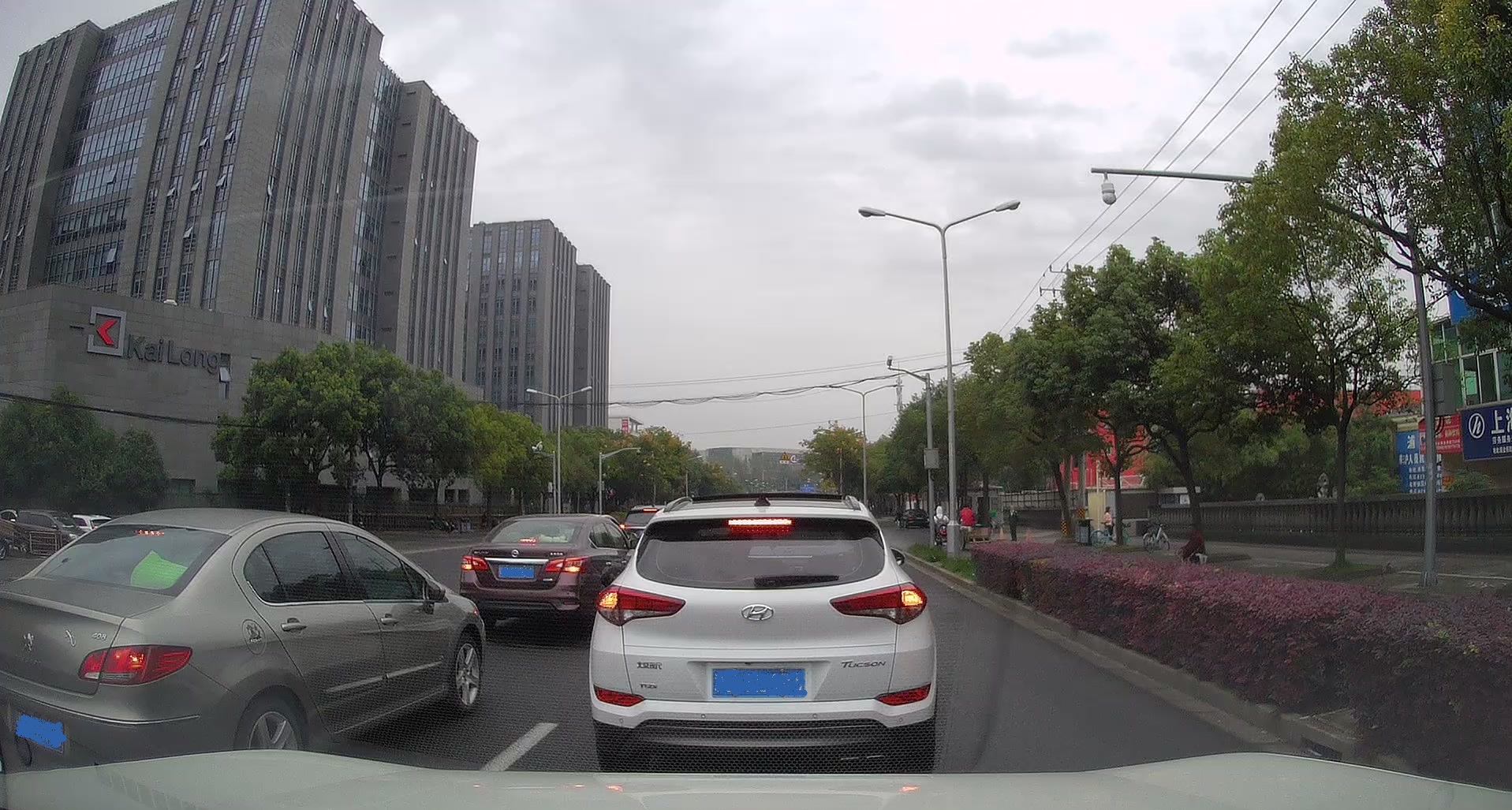}\includegraphics[width=0.33\linewidth]{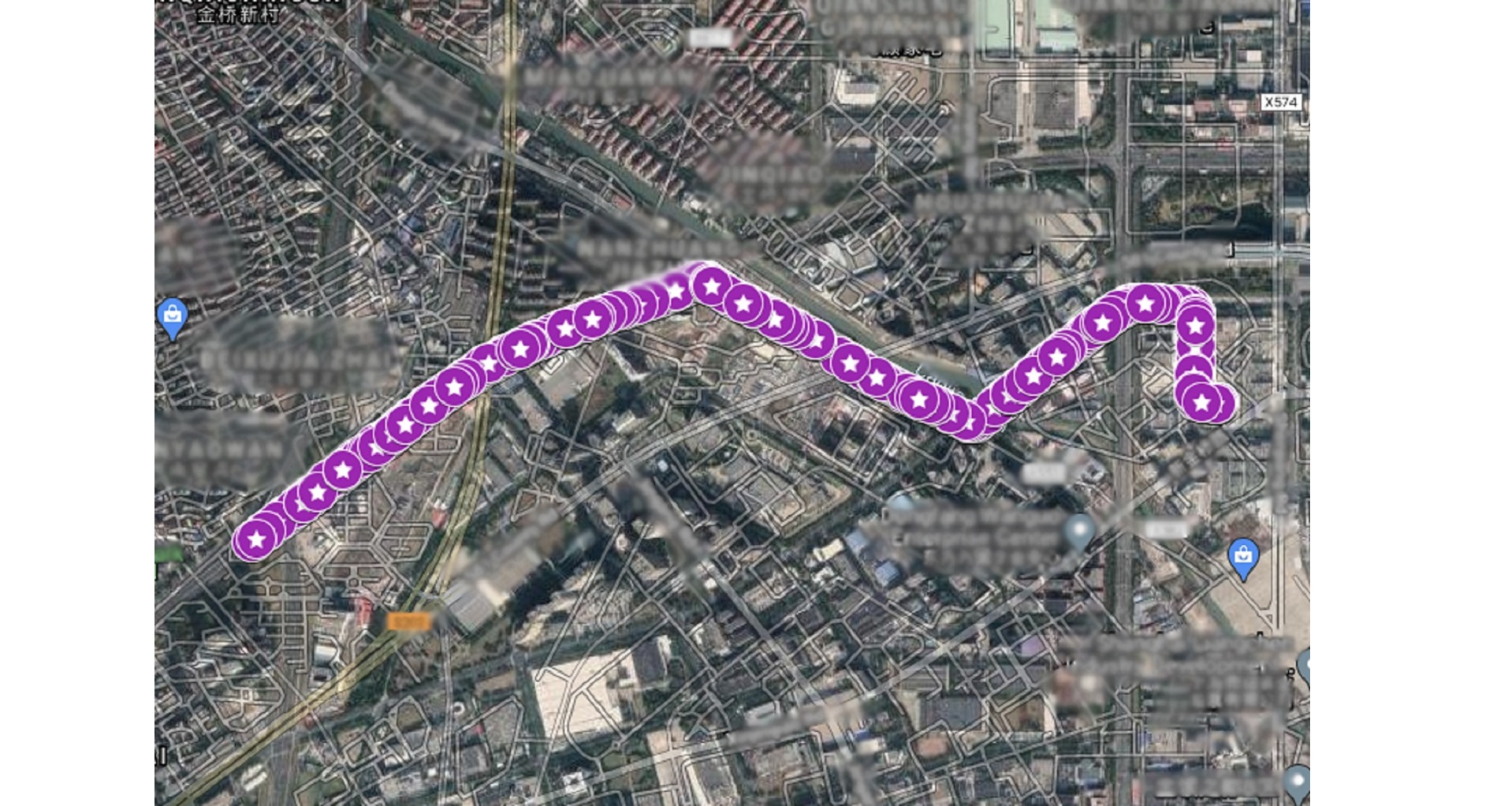}
   
   \caption{\textbf{Samples from Paris (top) and Shanghai (bottom) datasets.}\label{dataset_samples}}
\end{figure}

Our interest is to use CoordiNet for localization in large environments for practical applications. Publicly available benchmarks are limited to particular use case scenarios where data could be collected within few days. In scenarios related to development of localization functions for self-driving cars, one could rely on much larger scope of data available. In this section, we explore how well CoordiNet performs scales given amount of data provided for training is an order of magnitude higher compared to public benchmarks.

To do so, we use Oxford RobotCar dataset with a larger experiment : 15 sequences in training data resulting in 890k images in training set, 2 sequences used as validation and 7 test sequences, including the benchmark test sequence.

We also collected videos in Paris and Shanghai areas using dashcam cameras. We recovered ground truth poses from videos thanks to a large scale 3D reconstruction pipeline based on colmap SfM software~\cite{schoenberger2016sfm}, similar to~\cite{largescalesfm}. We provide examples from these datasets in figure~\ref{dataset_samples} and we briefly introduce them:

\begin{itemize}
    \item \textbf{Shanghai dataset}: a road of 3.3km, first part is highway and second part is smaller roads. Very busy traffic observed in most of sequences. Training set: 12 sequences with 123k images. Validation set: 6 sequences with 32k images. Test set: 9 sequences with 48k images.
    \item \textbf{Paris dataset}: A loop of 1.9km in urban area. One part along the Seine is challenging because of mirror reflection in buildings, the other goes through complex intersections with sometimes busy traffic. Training set: 6 sequences with 148k images. Validation set: 2 sequences with 30k images. Test set: 2 sequences with 13k images.
\end{itemize}

We train CoordiNet on these 2 datasets in addition to Oxford, but also use available implementation of Huang et al.~\cite{RVL} to compare performances with related methods (AD-PoseNet) and report results in table~\ref{tab:large_scale_results}.

\begin{table}[ht]
\footnotesize
\centering
\caption{Results on large scale dataset\label{tab:large_scale_results}.}
\begin{tabular}{|c|l|l|l|}
\hline
\multicolumn{2}{|c|}{\textbf{}}            & \textbf{CoordiNet (ours)} & \textbf{AD - PoseNet~\cite{RVL}} \\ \hline
\multirow{2}{*}{\textbf{Oxford}}  & median & \textbf{1.53m / 0.46°}  & 7.91m / 1.13°          \\ \cline{2-4} 
                                  & mean   & \textbf{7.11m / 2.93°}  & 19.89m / 4.51°                      \\ \hline
\multirow{2}{*}{\textbf{Shanghai}} & median & \textbf{0.69m / 0.69°}    & 9.24m / 0.47°         \\ \cline{2-4} 
                                  & mean   & \textbf{0.90m / 0.87°}    & 11.78m / 1.49°        \\ \hline
\multirow{2}{*}{\textbf{Paris}} & median & \textbf{0.29m / 0.29°}    & 3.75m / 1.03°         \\ \cline{2-4} 
                                  & mean   & \textbf{0.51m / 0.44°}    & 5.11m / 1.25°         \\ \hline
\end{tabular}
\end{table}

We see that CoordiNet outperforms previous SOTA pose regressor on large areas by an order of magnitude and observe that larger training sets allow to reach sub-meter accuracy on test data. Enlarged Oxford training set from 2 to 15 sequences enables to decrease mean error from 9.56m to 1.94m and median error from 3.55m to 1.25m on the same test sequence. We conclude that by collecting large image datasets and using CoordiNet as a pose regressor enables to reach reliable enough localization accuracy for selected practical applications. 

\subsection{Evaluation of localization under uncertainty}
\label{loc_uncertainty_exp}

\begin{figure}[!t]
   \centering
   \includegraphics[width=0.8\linewidth]{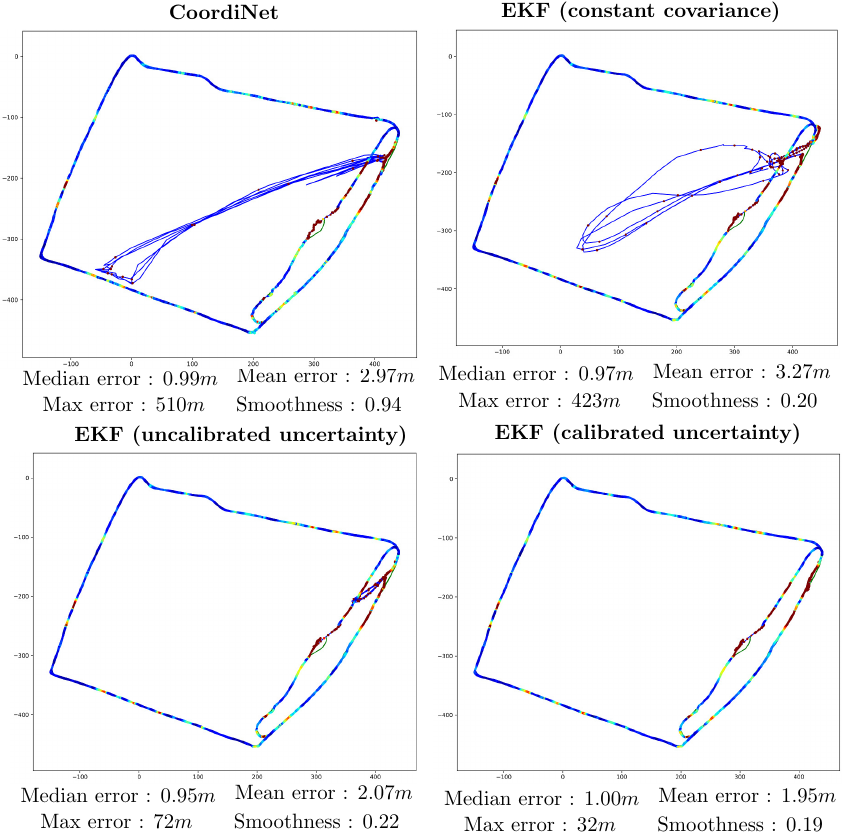}
   \caption{\textbf{Localization with uncertainty.} Top lane from left to right: CoordiNet predictions, EKF with CoordiNet poses and fixed covariance. Bottom lane from left to right: EKF with CoordiNet raw uncertainty, EKF with CoordiNet calibrated uncertainty. Colormap is the same as in figure~\ref{oxford_trajs}.}\label{uncertainty_effect}
\end{figure}

Rather than the raw CNN results, we are interested in the performance of the method outlined in \ref{locunderuncertainty}, where poses and uncertainties are fused into an EKF. We use the implementation provided by the ROS \textbf{robot\_localization} package with default parameters. We demonstrate superiority of our learned uncertainty over a fixed baseline using three experiments:
\begin{itemize}
    \item EKF with fixed covariance values,
    \item EKF with non-calibrated CoordiNet covariance values,
    \item EKF with calibrated CoordiNet covariance values, see section~\ref{subsec:calibrated_covariance}.
\end{itemize}
We report the results of these experiments on a full run of the Paris dataset in Figure~\ref{uncertainty_effect}. We evaluate a smoothness score $s$ of the trajectory (the lower the better) by computing the norm of the difference between two consecutive unitary directional vectors:
$$s = \frac{1}{N-2}\sum_{t=0}^{N-2} \norm{ \frac{T_{t+2} - T_{t+1}}{\norm{T_{t+2} - T_{t+1}}} - \frac{T_{t+1} - T_{t}}{\norm{T_{t+1} - T_{t}}}}.$$ 
As expected, coupled with an EKF the final trajectory gets smoother and the maximum error on the run is reduced. By rejecting outliers, the EKF reduces most of the time the mean error compared to the raw poses. 
We also show that it is crucial to estimate good covariance values in order to obtain the optimal trade-off between accuracy and smoothness: Coordinet + EKF with calibrated covariance performs the best in this experiment compared to fixed covariance values and to the baseline version.

We show in figure~\ref{screenshots} estimated covariance values as well as filtered trajectory in different parts of the Paris dataset map. Notice the shape of the covariance ellipsoid: for instance with the absence of lateral road markings, the covariance grows along the lateral direction.

\begin{figure}[ht]
    \centering
   \includegraphics[width=\linewidth]{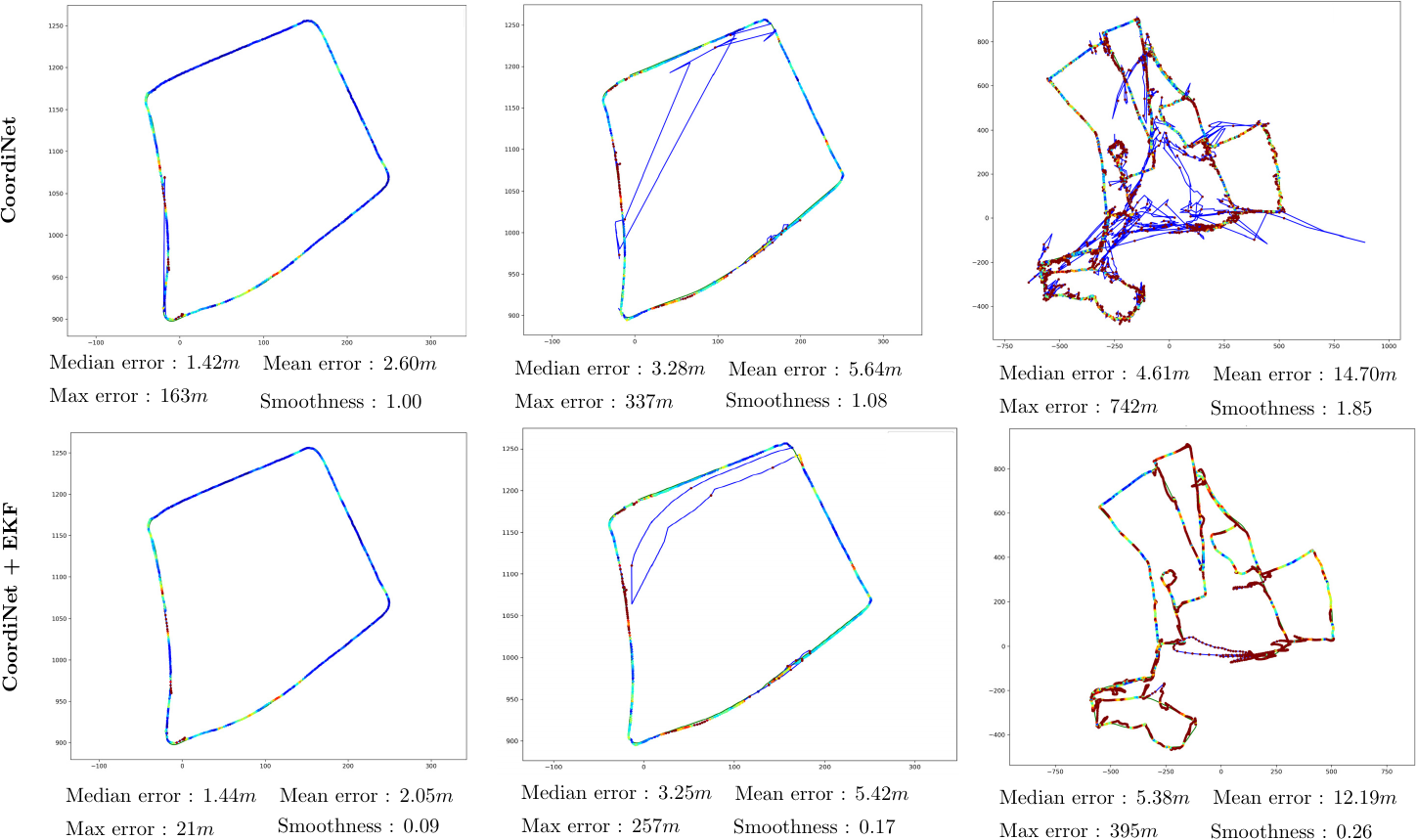}
   \caption{\textbf{EKF on Oxford experiment.} Top lane: CoordiNet trained with 2 runs. Bottom lane: EKF using CoordiNet poses and uncertainty. Colormap is the same as in figure~\ref{oxford_trajs}.}\label{fig:oxford_ekf}
\end{figure}

We also report the result on Oxford Robotcar experiment in figure~\ref{fig:oxford_ekf}. Again, the EKF is smoothing the trajectory and reduces the mean and maximum error on the overall trajectory. For this experiment, we use uncertainty without additional data. A careful reader should notice that reported results of CoordiNet in this table are slightly different than in table~\ref{tab:oxford_results}. This difference comes from the integration of Coordinet in our ROS framework. These results outperform methods with pose graph optimization reported in \cite{mapnet2018,RVL}.

\subsection{Ablation study}

Here, we evaluate all major contributions related to our method: coord convolution, loss function, pooling, geodesic rotation loss (\textbf{geo}) and the use of 3 translation losses instead of 1 (noted \textbf{split}). The proposed loss function with learned uncertainty is noted \textbf{heterosc.} for heteroscedatic uncertainty, we refer to usual "weighted pose loss" as \textbf{homosc}. Shanghai dataset (presented in \ref{large_scale_experiments}) is used for this experiment. \textit{EfficientNet b3} is used as image encoder

\begin{table}[ht]
\centering
\scriptsize
\caption{\label{tab:ablation_study_results} Ablation study on Shanghai dataset (errors in meters/degrees).}
\begin{tabular}{|l|l|l|l|l|l|l|}
\hline
\textbf{Loss} & \textbf{Coord} & \textbf{CWAP} & \textbf{Split} & \textbf{Rot} & \textbf{Median err.} & \textbf{Mean err} \\ \hline
heterosc.     & X              & X             & X              & geo.          & 0.58 / 0.20        & 1.26 / 0.36       \\ \hline
heterosc.     &                & X             & X              & geo.          & 0.69 / 0.69         & 0.90 / 0.87       \\ \hline
homosc.       &                & X             & X              & geo.          & 0.93 / 0.76         & 1.24 / 1.03       \\ \hline
Lt + Lr       &                & X             & X              & geo.          & 1.23 / 0.71         & 1.68 / 0.94       \\ \hline
heterosc.     &                &               & X              & geo.          & 0.95 / 0.6         & 1.18 / 0.87       \\ \hline
heterosc.     &                & X             &                & geo.          & 0.85 / 0.67         & 1.19 / 0.88       \\ \hline
heterosc.     &                & X             & X              & L1           & 0.74 / 0.75         & 0.94 / 1.25       \\ \hline
\end{tabular}
\end{table}

On this dataset, training with heteroscedatic uncertainty improve results by 16\%, coord convolutions by 16\%, confidence-weighted average pooling by 27\%, geodesic rotation loss by 30\% on mean rotation error and splitting the translation has a positive effect too.

\section{CONCLUSIONS}
\label{sec:conclusion}

In this paper, we proposed CoordiNet, a new deep neural network approach that pushes direct camera pose regression model accuracy one step further. Moreover, thanks to uncertainty quantification and large training sets, we demonstrated that our proposal can be integrated in real-time vehicle localization systems for an accurate pose estimation in large and busy urban environments.
This method could be improved in multiple ways: for instance by using sequences of images and more sophisticated uncertainty representation. Another challenge is to extrapolate to poses outside of the training distribution.

{\small
\bibliographystyle{ieee_fullname}
\bibliography{egbib}
}

\end{document}